\documentclass{llncs}
\usepackage{amsmath} 
\usepackage{amssymb} 
\usepackage{graphicx} 
\usepackage{color} 
\usepackage{amssymb} 
\usepackage[compatibility=false]{caption}
\usepackage{subcaption} 
\usepackage{verbatim}

\graphicspath{{./images/}}

\newcommand{\norm}[1]{\left\lVert#1\right\rVert} 
\newcommand{\inner}[2]{\left\langle #1,#2 \right\rangle}
\newcommand{\diag}{\mathop{\mathrm{diag}}}

\begin{document}
\title{A Large Deformation Diffeomorphic Approach to Registration of CLARITY Images via Mutual Information}
\author{Kwame S. Kutten\inst{1}
\and Nicolas Charon \inst{1} 
\and Michael I. Miller \inst{1} 
\and J. Tilak Ratnanather \inst{1} 
\and Jordan Matelsky \inst{1} 
\and Alexander D. Baden \inst{1} 
\and Kunal Lillaney \inst{1} 
\and Karl Deisseroth \inst{2} 
\and Li Ye \inst{2} 
\and Joshua T. Vogelstein \inst{1}}


\institute{Johns Hopkins University, Baltimore MD, USA 
\and
Stanford University, Stanford, CA, USA}

\maketitle
\begin{abstract}
CLARITY is a method for converting biological tissues into translucent and porous hydrogel-tissue hybrids.
This facilitates interrogation with light sheet microscopy and penetration of molecular probes while avoiding physical slicing.
In this work, we develop a pipeline for registering CLARIfied mouse brains to an annotated brain atlas.
Due to the novelty of this microscopy technique it is impractical to use absolute intensity values to align these images to existing standard atlases.
Thus we adopt a large deformation diffeomorphic approach for registering images via mutual information matching.
Furthermore we show how a cascaded multi-resolution approach can improve registration quality while reducing algorithm run time.
As acquired image volumes were over a terabyte in size, they were far too large for work on personal computers.
Therefore the NeuroData computational infrastructure was deployed for multi-resolution storage and visualization of these images and aligned annotations on the web.
\end{abstract}

\section{Introduction}
One of the most exciting recent advances in brain mapping is the introduction of CLARITY.
All cells are surrounded by a phospholipid bilayer which scatters light, rendering most biological tissues opaque to the naked eye.
Thus to use light microscopy, it is often necessary to physically slice brains.
Sectioning tissue has two major drawbacks for researchers interested in building whole brain connectomes.
First, slicing can dislocate synapses and axons necessary for tracing neuronal circuitry.
Second, the inter-sectional resolution will always be much lower than the intra-sectional resolution, making neurite tracing difficult~\cite{Kim}.
CLARITY avoids these problems by converting biological tissues into translucent and porous hydrogel-tissue hybrids.
This permits the penetration of photons and molecular probes while enabling interrogation using light sheet microscopy~\cite{Kim}.

CLARITY-optimized light sheet microscopy (COLM) was introduced to speed up image acquisition while maintaining high resolutions \cite{Tomer}.
Manipulation of these images is impossible on a desktop computer since each is over a terabyte in size.
Hence these images were ingested into the NeuroData computational cluster, which was designed for multi-resolution storage, access and visualization of large images \cite{Burns13,Kutten16b}.

Deformable registration of acquired images to a standard atlas is an essential step in building connectomes.
It is necessary for determining which brain regions axons pass through or the locations of synapses.
In this work we use a large deformation diffeomorphic technique for deformably registering CLARITY images to the Allen Reference Atlas (ARA).
Since its introduction in 2004, the ARA has been widely used by researchers to study brain anatomy, function and disease~\cite{Jones}.
As ARA images greatly differ from COLM volumes in appearance, we adopt Mutual Information (MI) matching during deformable registration.
We then apply this to the deformable registration of eleven COLM-acquired mouse brain images to the ARA.

\section{Image registration in the LDDMM framework}
The problem of deformable image registration is as follows.
Let $\Omega \subset \mathbb{R}^N$ be the background space where $N$ is the number of dimensions. Given template image $I_0: \Omega \to \mathbb{R}$ and target image $J_1: \Omega \to \mathbb{R}$ we seek a nonlinear map $\varphi$ such that $I_0 \circ \varphi^{-1}$ is aligned to $J_1$.
In biological imaging, deformations need to account for a large variety of local morphological variations.
Hence $\varphi$ should ideally be modeled as a diffeomorphism, i.e. a differentiable coordinate transform with differentiable inverse. 

\emph{Large Deformation Diffeomorphic Metric Mapping (LDDMM)} was introduced by Beg \textit{et al.} to compute these types of maps between images~\cite{Beg05}.
In LDDMM, time-varying velocity field $v: [0,1] \times \Omega \rightarrow \mathbb{R}^N$ flows $I_0$ to the space of $J_1$ over time domain $[0,1]$.
Diffeomorphic map $\phi_{st}: \Omega \to \Omega$ represents the coordinate transform from time $s \in [0,1]$ to time $t \in [0,1]$ where $s < t$.
The flow is defined by $\frac{d}{dt} \phi_{st} = v(t, \phi_{st})$ or in integrated form as $\phi_{st} = id + \int_s^t v(\tau,\phi_{s \tau}) d\tau$ with $id(x) \doteq x$.
Let the deformed template at time $t$ be defined by $I(t) = I_0 \circ \phi_{t0}$.
LDDMM finds optimal $v$ which minimizes the functional
\begin{equation}
 E(v) = R(v) + \frac{1}{2 \sigma^2} M(I(1), J_1)
 \label{eq:functional}
\end{equation}
where $M(I(1),J_1)$ is a matching term that is minimized when deformed template $I(1)$ is aligned with target $J_1$.
In Beg \textit{et al.}, the Sum of Squared Differences (SSD), $M(I(1),J_1) = \norm{I(1) - J_1}_{L^2}^2$, was adopted.

Regularization term $R(v) = \frac{1}{2}\int_0^1 \norm{Lv(t)}_{L^2}^2 dt$ has differential operator $L = \diag(L_1, \ldots, L_N)$ with identical entries $L_i = -\alpha \nabla^2 + \gamma$.
This Laplacian-based operator ensures that $v$ is smooth by penalizing second order derivatives of $v$.
Constant $\alpha >0$ determines the smoothness of the transform with higher $\alpha$-values yielding smoother transforms.
The constant  $\sigma >0$ determines the weight of the matching term relative to the regularization term and its chosen value typically depends on the level of noise in the image.

\section{Mutual Information approach for LDDMM}
As SSD is based on image subtraction it assumes that bright regions should be aligned to bright regions.
This assumption is routinely violated in microscopy where a wide variety of stains and fluorescent labels can be used to generate images that vary greatly in appearance. 
Hence ARA atlas image to CLARITY registration using SSD matching has been shown to give poor results.
A previously proposed workaround to this problem was to register the binary mask of the subject's brain to that of the atlas brain under SSD matching \cite{Kutten16b}.
In this ``Mask-LDDMM'' method, only edge information was incorporated and gray level values within the images were ignored.
While this method could accurately align superficial cortical structures its practical application was limited due to misalignment of deeper brain structures.

A more robust way to address this problem is to adopt Mutual Information (MI) as the matching term in LDDMM.
Since MI does not explicitly depend on grayscale values, it can be used to align corresponding image regions regardless of whether they share intensity values~\cite{Pluim}.
In this approach, we define the matching term as the negative MI
\begin{equation}
 M(I(1),J_1) = - \int_{-\infty}^{\infty} \int_{-\infty}^{\infty}  p_{IJ}(\eta,\xi) \log \left(\frac{p_{IJ}(\eta,\xi)}{p_I(\eta)p_J(\xi)}\right) d\eta d\xi \label{eq:mi}
\end{equation}
where $\eta \in \mathbb{R}$ and $\xi \in \mathbb{R}$ are intensity values from $I(1)$ and $J_1$ respectively.
Distributions $p_I(\eta)$, $p_J(\xi)$ and $p_{IJ}(\eta,\xi)$ come from the deformed template, target and joint histograms of the images.

Many past works have looked into combining various registration models with MI.
In \cite{Avants07} for example, the authors consider similar large deformations with MI but impose symmetric registration constraints between template and target.
For the applications of this paper, we are interested in template-to-target registration for which such a constraint is not necessary.
This can be done very conveniently within the standard LDDMM framework through the optimal control viewpoint presented recently in \cite{Miller15}, as we succinctly derive below in the context of our applications.

With the previous notations, the optimal control problem in our case is
\begin{equation}
 v^* = \arg\min\limits_v \{E(v):\; \partial_t I =  - \nabla I \cdot v,\; I(0) = I_0 \}     
 \label{eq:problem}
\end{equation}
where the state is the deformed template image $I$ and $v$ is the control that evolves the image through the advection equation $\partial_t I =  - \nabla I \cdot v$. 
The corresponding Hamiltonian $H(\rho,I,v) =  - \inner{\rho}{\nabla I \cdot v}_{L^2} - \frac{1}{2}\norm{Lv}_{L^2}^2$ is classically obtained by introducing a costate function $\rho$ in $L^2$ associated to the constraints.
The dynamics of optimal solutions to \eqref{eq:problem} can be then obtained from Pontryagin's Maximum Principle and are fully described by the associated Hamiltonian system.
Following \cite{Miller15} while keeping $M$ undefined we obtain the following system of equations:
\begin{subequations}
 \label{eq:solution}
 \begin{align}
  I(t) &= I_0 \circ \phi_{t0}  \label{eq:solution1}\\
  \rho(t) &= - \frac{1}{2 \sigma^2} \Big(\partial_I M\big(I(1),J_1\big) \circ \phi_{t1}\Big) |D\phi_{t1}| \label{eq:solution2}\\
  v(t) &= -(L^\dag L)^{-1}(\rho(t)\nabla I(t))  \label{eq:solution3}
 \end{align}
\end{subequations}

A notable advantage of this formulation is that the change in matching term $M$ only appears as the endpoint condition for the costate $\rho$ through the G\^{a}teaux derivative $\partial_I M\big(I(1),J_1\big)$.
In the case of MI matching term \eqref{eq:mi}, one has the derivative expression
\begin{equation}
 \partial_I M(I, J_1) = -\int_{-\infty}^{\infty} \int_{-\infty}^{\infty} \partial_I p_{IJ}(\eta,\xi) \log p_{J|I}(\xi|\eta) d\eta d\xi.
 \label{eq:miGradient}
\end{equation}
which can then be plugged in \eqref{eq:solution2}.

\section{Algorithm Implementation}
The set of equations in \eqref{eq:solution} do not entirely provide the solution to \eqref{eq:problem}.
The evolution of $\rho$ depends on the final state $I(1)$ which in turn depends on the velocity $v$ and thus $\rho$ as well. 
However, the solution itself can be obtained by an iterative fixed-point or equivalently gradient descent procedure using $\nabla_v E(t) = v(t) + (L^\dag L)^{-1}(\rho(t)\nabla I(t))$.

In our numerical implementation, we follow a similar discretization approach as Beg \textit{et al} \cite{Beg05}.
Time domain $[0,1]$ is discretized into $T$ uniformly spaced time-steps such that $0 = t_0 < t_1 < \ldots < t_{T-1} = 1$.
We initialize with $v(t_j) = 0$ and $I(t_j) = I_0$ for each time-step $j \in \{0, \ldots, T-1\}$.
In each iteration we find the new time-varying velocity for all time steps $j$, $v_\mathrm{new}(t_j) = v(t_j) - \varepsilon \nabla_v E(t_j)$ where $I(t_j)$, $\rho(t_j)$ are obtained from equations \eqref{eq:solution1} and \eqref{eq:solution2}. 
Following \cite{Beg05}, maps of the form $\phi_{t_j 0}$ and $\phi_{t_j 1}$ are computed using semi-Lagrangian integrators while operators $L$ and $K$ are applied in the Fourier domain.
Starting from an initial step-size $\varepsilon_0$, a adaptive step rule is adopted to update $\varepsilon$ in each iteration.

We follow Mattes \emph{et al.} \cite{Mattes} to evaluate MI equation \eqref{eq:mi} and its gradient \eqref{eq:miGradient}.
A Parzen window approach is used to estimate the joint distribution $p_{IJ}$ with 3rd degree B-splines for the template intensities and 0th order B-splines on the target intensities.
Numerically, Mattes Mutual Information and its derivatives were computed from its implementations in the open source Insight Segmentation and Registration Toolkit (ITK) \cite{Avants12}.
Our code also takes advantage of ITK's virtual domain framework, enabling evaluations of MI values and derivatives at a lower resolution than the template and target images \cite{ITK2}.
Using SimpleITK \cite{Lowekamp}, we packaged this implementation into the \emph{NeuroData Registration (ndreg)} Python module \cite{Kutten16b} which was designed for alignment of images stored in NeuroData infrastructure.
The \emph{NeuroData Input/Output} (ndio) module, a python implementation of the \emph{Connectome Annotation for Joint Analysis of Large data} (CAJAL) library \cite{GrayRoncal}, was also written for downloading and uploading image data stored in our computational framework.

\section{Results}
\subsection{MI Registration Pipeline}
\begin{figure}
 \centering
 \includegraphics[width=0.8\columnwidth]{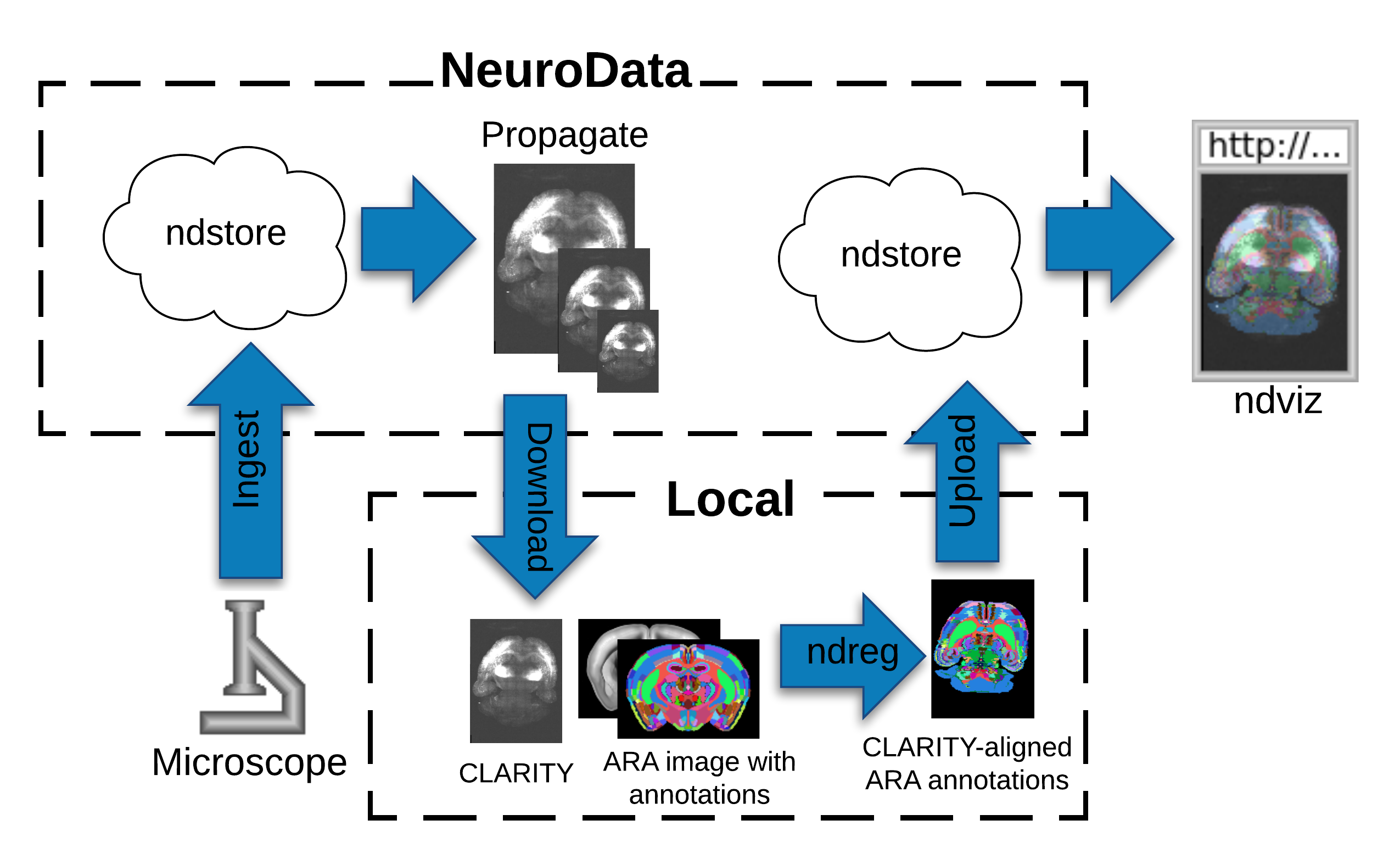}
 \caption{Pipeline from image acquisition with microscope to visualization on the web.
 Acquired image data is stitched and ingested into NeuroData Storage (ndstore).
 After propagation to lower resolutions each CLARITY volume was registered to the ARA with MI-LDDMM as implemented in ndreg.
 CLARITY-aligned ARA annotations are uploaded to ndstore where they can be visualized over the terabyte-scale acquired volume.}
 \label{fig:workflow}
\end{figure}

Eleven mouse brains were CLARIfied and imaged with COLM as described by Tomer \textit{et al.}~\cite{Tomer}
Acquired image data was then stitched and ingested into the NeuroData cluster.
At their full resolution of 0.6 $\mu m$ x 0.6 $\mu m$ x 6 $\mu m$, each volume was over a terabyte in size.
Image data was then propagated to more manageable lower resolutions (Fig~\ref{fig:workflow}).

\begin{figure}
 \centering
 \includegraphics[width=1.0\columnwidth]{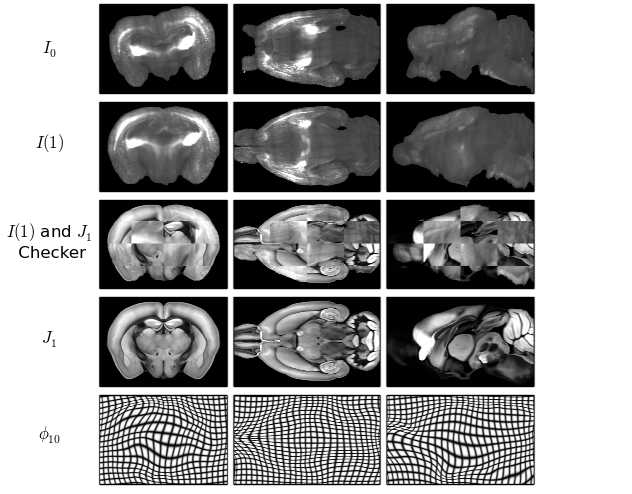}
 \caption{Detailed MI-LDDMM registration results for a CLARITY mouse brain.
 First, second and third columns are coronal, axial and sagittal slices respectively.
 First row and forth row are CLARITY template and ARA target respectively.
 Second and third rows are the deformed template and its checkerboard pattern with the ARA respectively.
 Final row is the deformation grid.}
 \label{fig:clairitySlices}
\end{figure}

\begin{figure}
 \centering
 \begin{subfigure}{0.18\columnwidth}
  \captionsetup{justification=centering} 
  \includegraphics[width=\textwidth]{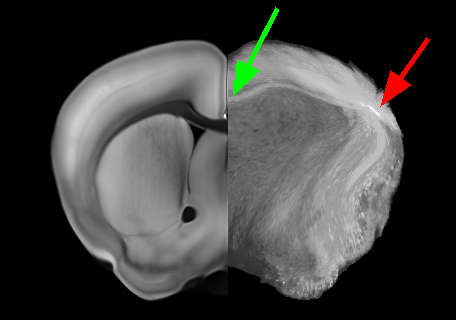}  
  \caption{SSD- LDDMM}
  \label{fig:clarityCoronalSSD}
 \end{subfigure}
 \begin{subfigure}{0.18\columnwidth}
  \captionsetup{justification=centering} 
  \includegraphics[width=\textwidth]{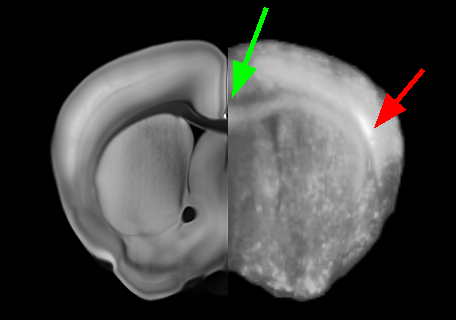}  
  \caption{Mask- LDDMM}
  \label{fig:clarityCoronalMask}
 \end{subfigure}
 \begin{subfigure}{0.18\columnwidth}
  \captionsetup{justification=centering} 
  \includegraphics[width=\textwidth]{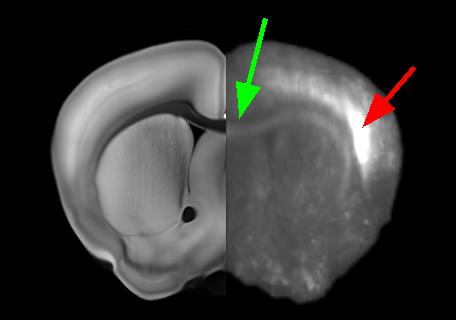}  
  \caption{MI- LDDMM}
  \label{fig:clarityCoronalMI}
 \end{subfigure}
 \begin{subfigure}{0.18\columnwidth}
  \captionsetup{justification=centering} 
  \includegraphics[width=\textwidth]{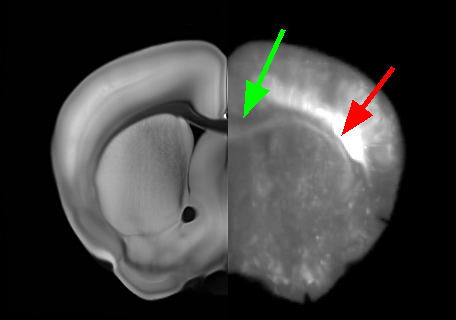}  
  \caption{SyN \\ \mbox{ }}
  \label{fig:clarityCoronalSyN}
 \end{subfigure}
 \begin{subfigure}{0.2\columnwidth}
  \captionsetup{justification=centering} 
  \includegraphics[width=\textwidth]{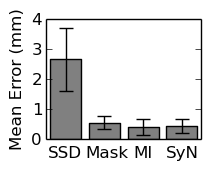}  
  \caption{Landmark Error}
  \label{fig:lmkError}
 \end{subfigure}
 \caption{Comparison of SSD-LDDMM (\subref{fig:clarityCoronalSSD}), Mask-LDDMM (\subref{fig:clarityCoronalMask}), MI-LDDMM (\subref{fig:clarityCoronalMI}) and SyN ANTs (\subref{fig:clarityCoronalSyN}) registration of CLARITY volume.
  Panes (\subref{fig:clarityCoronalSSD}-\subref{fig:clarityCoronalSyN}) have an ARA coronal slice on the left juxtaposed to the corresponding aligned CLARITY slice on the right.
  Green arrows point out that the corpus callosum is misaligned by SSD-LDDMM but aligned correctly by MI matching.
  Red arrows show that SSD-LDDMM distorts bright regions.
  Fiducial landmarks were manually placed in the corpus callosum, and midbrain of the acquired volumes.
  Pane (\subref{fig:lmkError}) compares mean errors between the deformed CLARITY and ARA landmarks after registration.
 }
 \label{fig:comparison}
\end{figure}

For registration, each CLARITY volume $I_0$ was downloaded from the NeuroData cluster.
They were resampled to a 50 $\mu m$ resolution and registered to ARA atlas image $J_1$ with 12-parameter affine alignment under MI matching.
Deformable registration was then done with MI-LDDMM as implemented in ndreg.
A cascaded-$\alpha$ approach was adopted in which a smoother registration with $\alpha=0.05$ was followed by registrations at $\alpha=0.02$ and $\alpha=0.01$ to refine the results (Fig.~\ref{fig:clairitySlices}).
Both ARA-aligned CLARITY volumes and CLARITY-aligned ARA annotations were uploaded to the NeuroData cluster.
This allowed us to visualize ARA annotations overlaid on the terabyte-scale CLARITY images (Fig~\ref{fig:workflow}).
Deformable registration was done with SSD-LDDMM and Mask-LDDMM to demonstrate the advantages of MI-LDDMM.
For validation MI-LDDMM was also compared to SyN ANTs \cite{Avants07} with MI cost and $\sigma=1.0$ mm Gaussian regularization (Fig.~\ref{fig:comparison}).

\subsection{Multi-resolution registration}
\begin{figure}
 \centering
 \includegraphics[width=0.8\columnwidth]{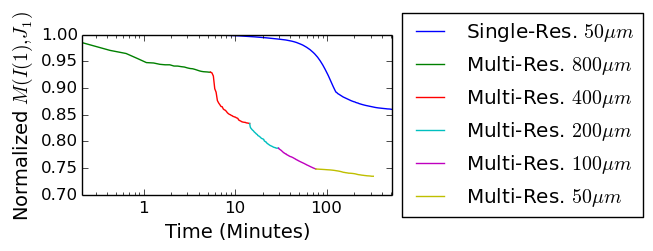}
 \caption{Comparison of the single and multi-resolution MI-LDDMM for one of the CLARITY volumes. Matching term $M(I(1), J_1)$ was normalized to range of [0,1] by ratio $\frac{M(I(1), J_1) - M(J_1, J_1)}{M(I_0, J_1) - M(J_1, J_1)}$, at each iteration}
 \label{fig:clarityMultiscalePlot}
\end{figure}

The typical run-time of MI-LDDMM can be particularly long. 
Thus we coupled our implementation with a cascaded multi-resolution approach where the optimization problem is first solved on a coarsened grid with ITK's virtual domain infrastructure.
Output vector fields $v$ are then interpolated to initialize optimization at the next higher resolution.
In this experiment, ARA registration was ran on all 11 CLARITY brains at 800 $\mu m$, 400 $\mu m$, 200 $\mu m$, 100 $\mu m$ and then 50 $\mu m$ resolutions with $\alpha=0.02$.
These results were compared to a single-resolution alignment at 50 $\mu m$.
In Figure~\ref{fig:clarityMultiscalePlot} it is clear that the multi-resolution optimization was more efficient than the single-resolution trial.
In this example, a decline below 0.9 for normalized $M(I(1), J_1)$ took over 100 minutes with single-resolution alignment and only 10 minutes with multi-resolution registration.
The multi-resolution registration was also more optimal as it terminated at a lower $M(I(1),J_1)$ value (Fig.~\ref{fig:clarityMultiscalePlot}).
This suggests that the multi-resolution method also prevents the algorithm from stopping at local minima of the functional.

\section{Conclusion}
In this work we implemented a large deformation diffeomorphic algorithm for registering images using mutual information matching.
We then applied it in a pipeline to register the ARA to CLARITY image volumes for overlay at the terabyte scale.
We also demonstrated how a multi-resolution approach can improve registration quality while reducing algorithm run time.

\section{Acknowledgments}
The authors are grateful for support from the DARPA SIMPLEX program through SPAWAR contract N66001-15-C-4041, DARPA GRAPHS N66001-14-1-4028.

\bibliographystyle{splncs03}
\bibliography{paper344}

\begin{thebibliography}{10}
\providecommand{\url}[1]{\texttt{#1}}
\providecommand{\urlprefix}{URL }

\bibitem{Avants07}
Avants, B.B., Epstein, C.L., Grossman, M., Gee, J.C.: Symmetric diffeomorphic
  image registration with cross-correlation: Evaluating automated labeling of
  elderly and neurodegenerative brain. Medical Image Analysis  12(1),  26--41
  (2007)

\bibitem{Avants12}
Avants, B.B., Tustison, N.J., Song, G., Wu, B., Stauffer, M., McCormick, M.M.,
  Johnson, H.J., Gee, J.C., {The Insight Software Consortium}: A unified image
  registration framework for {ITK}. In: Biomedical Image Registration, vol.
  7359, pp. 266--275. Springer-Verlag (2012)

\bibitem{Beg05}
Beg, M.F., Miller, M.I., Trouv\'e, A., Younes, L.: Computing large deformation
  metric mappings via geodesic flows of diffeomorphisms. International Journal
  of Computer Vision  61(2),  139--157 (2005)

\bibitem{Burns13}
Burns, R., Roncal, W.G., Kleissas, D., Lillaney, K., Manavalan, P., Perlman,
  E., Berger, D.R., Bock, D.D., Chung, K., Grosenick, L., Kasthuri, N., Weiler,
  N.C., Deisseroth, K., Kazhdan, M., Lichtman, J., Reid, R.C., Smith, S.J.,
  Szalay, A.S., Vogelstein, J.T., Vogelstein, R.J.: The {Open Connectome
  Project} data cluster: Scalable analysis and vision for high-throughput
  neuroscience. Proceedings of the 25th International Conference on Scientific
  and Statistical Database Management  (2013)

\bibitem{ITK2}
Johnson, H.J., McCormick, M.M., Ib{\'a}{\~n}ez, L.: The ITK Software Guide Book
  2: Design and Functionality. The Insight Software Consortium, 4 edn. (July
  2016)

\bibitem{Jones}
Jones, A.R., Overly, C.C., Sunkin, S.M.: The {Allen Brain Atlas}: 5 years and
  beyond. Nature Reviews Neuroscience  10,  821--828 (2009)

\bibitem{Kim}
Kim, S.Y., Chung, K., Deisseroth, K.: Light microscopy mapping of connections
  in the intact brain. Trends in Cognitive Sciences  17(12),  596--599 (2013)

\bibitem{Kutten16b}
Kutten, K.S., Vogelstein, J.T., Charon, N., Ye, L., Deisseroth, K., Miller,
  M.I.: Deformably registering and annotating whole {CLARITY} brains to an
  atlas via masked {LDDMM}. In: Proc SPIE 9896: Optics, Photonics and Digital
  Technologies for Imaging Applications IV (2016)

\bibitem{Lowekamp}
Lowekamp, B.C., Chen, D.T., Ib{\'a}{\~n}ez, L., Blezek, D.: The design of
  {SimpleITK}. Frontiers in Neuroinformatics  7 (2013)

\bibitem{Mattes}
Mattes, D., Haynor, D.R., Vesselle, H., Lewellen, T.K., Eubank, W.: Nonrigid
  multimodality image registration. In: Sonka, M., Hanson, K.M. (eds.) Proc.
  SPIE 4322, Medical Imaging: Image Processing. pp. 1609--1620 (2001)

\bibitem{Miller15}
Miller, M.I., Trouv{\'e}, A., Younes, L.: Hamiltonian systems and optimal
  control in computational anatomy: 100 years since {D'Arcy Thompson}. Annual
  Review of Biomedical Engineering  17,  447--509 (2015)

\bibitem{Pluim}
Pluim, J.P.W., Maintz, A., Viergever, M.A.: Mutual information based
  registration of medical images: a survey. IEEE Transactions on Medical
  Imaging  22(8),  986--1004 (2003)

\bibitem{GrayRoncal}
Roncal, W.R.G., Kleissas, D.M., Vogelstein, J.T., Manavalan, P., Lillaney, K.,
  Pekala, M., Burns, R., Vogelstein, R.J., Priebe, C.E., Chevillet, M.A.,
  Hager, G.D.: An automated images-to-graphs framework for high resolution
  connectomics. Frontiers in Neuroinformatics  9 (2014)

\bibitem{Tomer}
Tomer, R., Ye, L., Hsueh, B., Deisseroth, K.: Advanced {CLARITY} for rapid and
  high-resolution imaging of intact tissues. Nature Protocols  9(7),
  1682--1697 (2014)

\end{thebibliography}
 
\end{document}